\documentclass[runningheads,a4paper]{llncs}
\usepackage{wrapfig}
\usepackage{amsmath}
\usepackage{makecell}
\usepackage{tabularx} 
\usepackage{hyperref}
\usepackage{booktabs}
\usepackage{times}
\usepackage{amssymb}
\usepackage{graphicx}
\usepackage{multirow}
\usepackage{xcolor}%
\usepackage{url}
\usepackage{float}
\usepackage{caption}
\usepackage{subcaption}
\newcommand{\keywords}[1]{\par\addvspace\baselineskip
\noindent\keywordname\enspace\ignorespaces#1}

\urldef{\mailsa}\path||

\urldef{\mailsb}\path| |

\begin{document}

\title{Scale-free Characteristics of Multilingual Legal Texts\\ and the Limitations of LLMs}

\titlerunning{Scale-Free Characteristics of Multilingual Legal Texts and the Limitations of LLMs}

\author{Haoyang   Chen \and Kumiko Tanaka-Ishii  }


\authorrunning{Haoyang   Chen \and Kumiko Tanaka-Ishii}

\institute{Faculty of Science and Engineering, Waseda University,\\ 169-8555, 3-4-1 Okubo, Shinjuku, Tokyo, Japan \\
\url{https://ml-waseda.jp/} \\
\mailsa\\
\mailsb
}

\index{Chen, Haoyang}
\index{Tanaka-ishii, Kumiko}

\toctitle{} \tocauthor{}

\maketitle

\begin{abstract}
We present a comparative analysis of text complexity across domains using scale-free metrics. We quantify linguistic complexity via Heaps' exponent $\beta$ (vocabulary growth), Taylor's exponent $\alpha$ (word-frequency fluctuation scaling), compression rate $r$ (redundancy), and entropy. Our corpora span three domains: legal documents (statutes, cases, deeds) as a specialized domain, general natural language texts (literature, Wikipedia), and AI-generated (GPT) text. We find that legal texts exhibit slower vocabulary growth (lower $\beta$) and higher term consistency (higher $\alpha$) than general texts. Within legal domain, statutory codes have the lowest $\beta$ and highest $\alpha$, reflecting strict drafting conventions, while cases and deeds show higher $\beta$ and lower $\alpha$. In contrast, GPT-generated text shows the statistics more aligning with general language patterns. These results demonstrate that legal texts exhibit domain-specific structures and complexities, which current generative models do not fully replicate.

\end{abstract}

\section{Introduction}
\vspace{-0.2cm}
\label{sec:intro}
With the advent of AI, legal texts have become an important challenge in the fields of AI and NLP. Legal texts exhibit probabilistic patterns as in the natural language~\cite{shannon1948mathematical,coverthomas}, but there is more. Above all, legal texts are considered difficult because they employ specialized vocabulary and follow rigorous drafting conventions~\cite{bommarito2010mathematical,katz2014measuring}, and legal discourse must be consistent in its content to adjudicate human behavior \cite{Loevinger1963JurimetricsTM,de2010jurimetrics}.

In this work, we consider how such characteristics of legal text can be quantified by complexity measures. Our analysis reveals that legal texts exhibit distinctive statistical patterns: specifically, they have slower vocabulary growth (smaller $\beta$) and greater term consistency (larger $\alpha$) than general-language texts. In contrast, GPT-generated texts show higher $\beta$ and lower $\alpha$, aligning more closely with general language patterns. 

On the basis of scale-free metrics, we consider the gzip-based compression rate $r$. Previous studies have shown that $r$ follows a power law scaling with text size (e.g., \cite{springer21}) and that legal drafting yields higher $r$ than literary prose \cite{friedrich2021complexity}. In this study, we include this compression rate and provide evidence of both its effectiveness and its limitations. 

The perspective of this study differs from previous substantial works on categorization and clustering, which use supervised or unsupervised methods that use classification accuracy to distinguish legal from general text, but lack an explicit mathematical rationale for the underlying differences. Instead, we show that the key divergence between legal and general texts lies in their scale-free discourse structure. 

Understanding these complexity signatures is crucial for processing legal texts, as the current AIs themselves exhibit scale-free behavior \cite{kaplan}, but we show that they are limited in their capability of capturing legal drafting conventions. We argue that our findings must motivate a fundamental reconsideration of processing strategies and provide a road map for analyzing other challenging text genres.

\section{Related Work}
\vspace{-0.2cm}
\label{sec:related work}
In this section, we review approaches to analyze textual data. We then contrast with our framework, {emphasizing how our method is different by directly quantifying linguistic complexity rather than relying on classification accuracy or black-box methods.}

\subsection{Text Categorization and Clustering Methods}
\vspace{-0.2cm}
{Traditional approaches to textual analysis frequently utilize machine learning techniques, such as support vector machines (SVM), k-means clustering, and transformer-based embeddings, to categorize or cluster documents by topic, style, or content type} \cite{Sebastiani2002Text,Manning1999Foundations,Blei2003Latent,Reimers2019Sentence}. Although these methods achieve high classification accuracy, they operate as `black boxes' and do not reveal the underlying linguistic principles, such as vocabulary growth dynamics or term clustering, that distinguish domains. In contrast, our scale-free framework directly measures statistical exponents on texts, {thus quantifying underlying structural characteristics and linguistic complexities across text domains.}

\subsection{Machine Learning and LLMs in Text Analysis}
\vspace{-0.2cm}
Machine learning, growing from early rule-based systems to current deep neural architectures, has revolutionized text processing. Transformer-based language models like GPT-4 \cite{openai2023gpt4} and Llama \cite{touvron2023llama} demonstrate state-of-the-art performance in generation and understanding tasks. However, capturing the specialized structure of domain-specific texts remains challenging \cite{chalkifair,Deroy2024Applicability,surveylaw}. Here, our approach using scaling-law metrics shows that, although modern models can produce fluent text, they often mirror general language statistics rather than the precise patterns of specialized corpora.

\subsection{Legal Texts and AI}
\vspace{-0.2cm}
Efforts to adapt AI to legal language include domain-specific pre-training (e.g., Legal-BERT~\cite{chalkidis-etal-2020-legal} on statutes, opinions, and contracts) and generic models on benchmarks such as LexGLUE \cite{chalkidis-etal-2022-lexglue}. Extensions of neural network architectures, such as Longformer, can tackle ultra-long documents, while multi-label classification further refines performance \cite{beltagy2020longformerlongdocumenttransformer,chalkidis-etal-2021-multieurlex}. {However, these methods primarily measure success through accuracy in downstream tasks and treat legal texts as classification targets rather than analyzing their linguistic complexity. By contrast, our approach explicitly quantifies textual characteristics— vocabulary growth, contextual clustering, and redundancy—thereby uncovering statistical signatures inherent in legal language that conventional AI approaches typically overlook.}

{To summarize, our scale-free framework measures the underlying linguistic structure of texts. Unlike existing approaches that rely on accuracy and model optimization, our method directly quantifies the complexity of texts, providing deeper insights into the structural properties of specialized texts such as legal documents.}
\vspace{-0.2cm}
\section{Methodology}
\vspace{-0.2cm}
\label{sec3}
This section describes the methodology used to assess the differences in complexity between the text categories. We propose two types of metrics: scale-free properties (our original focus) and general metrics (compression rate), which have frequently been used in previous work on text complexity~\cite{katz2014measuring,entropy16,friedrich2021complexity}.
\vspace{-0.2cm}
\subsection{Metrics Related to Scale-Free Properties}
\vspace{-0.2cm}
\label{sec:sec3.1}
Texts often exhibit universal statistical behaviors in the form of scale-free properties, meaning these properties hold regardless of text size. This scale-free quality is crucial because many statistical metrics, including entropy, typically depend on the corpus size~\cite{cl15}. A comprehensive summary of scale-free properties is given in \cite{springer21}, which categorizes them into frequency distribution and long-memory aspects. We will demonstrate that different text domains possess distinct characteristics in both of these aspects.

Regarding the frequency distribution, Zipf's law is the best-known power-law property of language, describing the relationship between word frequency rank and frequency. However, because the Zipf exponent is close to $-1$ for most natural texts, Zipf's law alone does not distinguish between corpora. We therefore focus on the type-token relationship (Heaps' law). Given a text $X$ with vocabulary size $V$ and total words $N$, we have:
\begin{equation}
V \propto N^{\beta}.
\end{equation}
The exponent $\beta$ is 1 for certain random texts, but for natural language it is typically smaller (often around 0.7)~\cite{springer21}. Later, we will show that specialized domain texts tend to have even smaller $\beta$ values, indicating slower vocabulary growth.

For long-memory properties, while traditional long-range correlation analysis can be problematic for text~\cite{springer21}, Taylor's law provides an alternative. We divide a text into segments and measure the mean $\mu_w$ and standard deviation $\sigma_w$ of occurrences of each word $w$. Across words, these follow:
\begin{equation}
\sigma \propto \mu^{\alpha}.
\end{equation}
Excluding token types with zero variance, we perform a linear regression and take the slope \(\alpha\) as the Taylor exponent \cite{taylor-nature}. Taylor's exponent $\alpha$ is 0.5 for a shuffled text, and $0.5<\alpha<1$ for natural texts~\cite{takahashi-tanaka-ishii-2019-evaluating}. We compute $\alpha$ in order to capture contextual clustering of words: a higher $\alpha$ indicates a greater context dependence. 

\subsection{Entropy and Compression Rate Analysis}
\vspace{-0.2cm}
\label{subsection3.1}
Information‐theoretic measures have long been used to investigate legal language complexity. Early work used Shannon entropy \cite{shannon1948mathematical} to distinguish legal from general‐language corpora \cite{katz2014measuring}. \cite{friedrich2021complexity} later proposed pairing a “normalized vocabulary entropy” with a gzip‐based compression rate to separate text types.

We also use the compression rate $r$, computed as the ratio of the original text length to the compressed text length, to providing an empirical measure of text redundancy:
\begin{equation}
r = \frac{|X|}{|{\rm h}(X)|},
\end{equation}
where $X$ is a text sample, $|X|$ is the length in bits of $X$, and ${\rm h}(X)$ is that of the compressed version. Apart from this compression rate, \cite{friedrich2021complexity} defined a new measure called `normalized entropy' as follows:
\begin{equation}
      H_{\mathrm{norm}} \;=\; \frac{H(X)}{\log_{2}\lvert V\rvert},
\end{equation}
\vspace{-0.2cm}
where, $H(X)$ is the Shannon entropy:
\begin{equation}
    H(X) \;=\; -\sum P(w)\,\log_{2}P(w),   
\end{equation}
where $P(w)$ is the simple relative entropy. 

We will reserve discussion of this measure till Sec.~\ref{sec:fred}, where we show that, by itself, \(H_{\mathrm{norm}}\) does not effectively distinguish text types. In the following sections, our primary analysis focuses on the scale‐free exponents \(\beta\), \(\alpha\) (Sect.~\ref{sec:sec3.1}) and the compression rate \(r\).
\begin{figure}[b]
\vspace{-0.2cm}
\centering
\includegraphics[width=0.9\textwidth]{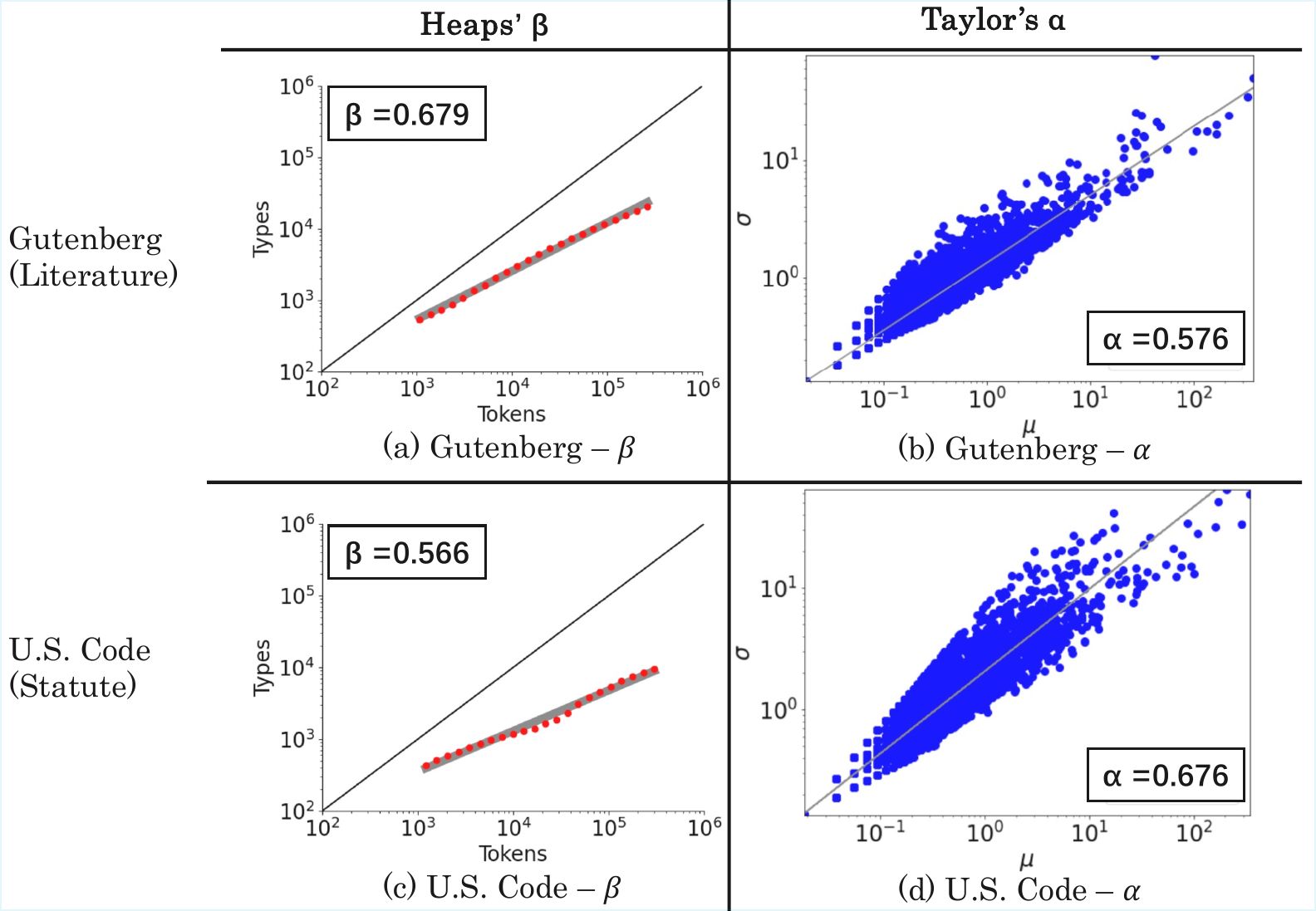}
\caption{Comparison of $\alpha$ (right) and $\beta$ (left) between literature and U.S. Code examples via Heaps' and Taylor's analyses.}
\label{fig:test}
\vspace{-0.5cm}
\end{figure}

\section{Illustrative Examples of Text Complexity}
\vspace{-0.2cm}

Before presenting our large-scale results, we illustrate the metrics in two examples. We selected one literature text (``Terminal Compromise'' by Winn Schwartau; 265,780 words) and one legal code text (the ``Abandoned Property Article'' from the U.S. Code; 300,770 words). Fig.~\ref{fig:test}(a) and Fig.~\ref{fig:test}(c) show Heaps' analysis for these texts. The U.S. Code example has a much smaller $\beta$ than the literature example, indicating slower vocabulary growth for that text. Similarly, Fig.~\ref{fig:test}(b) and Fig.~\ref{fig:test}(d) show Taylor's analysis: the literature example has a smaller $\alpha$, whereas the U.S. Code example's larger $\alpha$ indicates more clustered word usage. We also computed the compression rate for these two examples: U.S. Code ($r=4.383$) versus literature ($r=2.577$). The higher compression for the Code reflects its greater repetitiveness. In the following sections, we show that these tendencies generalize across our datasets.

\section{Data}
\begin{table}[t]
\caption{Basic statistics for different text categories.}
\centering
\begin{tabular}{|l|r|r|r|r|}
\hline 
Category & Files Count & \makecell[c]{Total Words\\($N$)} & \makecell[c]{Vocabulary\\($V$)} & Average Words/File \\ 
\hline 
\multicolumn{5}{|c|}{\textbf{Legal Texts (greenish in graphs)}} \\ 
\hline 
Fr\_Code (statute)               & 31  &  9,127,733 &   80,979 &     294,443.00 \\ 
USCode (statute)                 & 19  &192,630,688 &  250,466 &  10,138,457.26 \\ 
Australian\_corpus (statute)     & 39  & 11,585,900 &  182,871 &     297,074.36 \\ 
\hline 
Harvard\_Case\_Law (case)        & 28  & 59,591,032 &  398,401 &   2,128,251.14 \\ 
\hline 
Patent (deed)                    & 34  & 10,000,000 &  226,308 &     294,117.65 \\ 
LEDGAR (deed)                    & 47  & 14,225,102 &   40,621 &     302,661.74 \\ 
de\_contract (deed)              & 36  & 10,394,439 &  336,788 &     288,734.42 \\ 
\hline 
\multicolumn{5}{|c|}{\textbf{Non-legal Language (reddish in graphs)}} \\ 
\hline 
Literature                       & 612 &187,804,121 &1,791,761 &     306,869.48 \\ 
Wikipedia                        & 34  & 10,000,000 &  251,767 &     294,117.65 \\ 
Wikipedia (random)               & 54  & 16,201,949 &  574,704 &     300,036.09 \\ 
\hline 
\multicolumn{5}{|c|}{\textbf{GPT-Generated (blackish in graphs)}} \\ 
\hline 
chatgpt\_news                    & 24  &  7,036,150 &   86,541 &     293,172.92 \\ 
chatgpt\_paraphrase              & 35  & 10,500,000 &   63,562 &     300,000.00 \\ 
chatgpt\_caselaw                 & 30  &  8,697,428 &   43,532 &     289,914.27 \\ 
\hline 
\end{tabular}
\label{tab:basic_stats}
\vspace{-0.3cm}
\end{table}
Table~\ref{tab:basic_stats} summarizes the sizes and vocabularies of each dataset. We compiled publicly available corpora in three domains:
\begin{itemize}
\item \textbf{Legal texts}: Statutory law (French Code, U.S.\ Code, Australian legislation \cite{butler-2024-open-australian-legal-corpus}), cases (the Caselaw Access Project \cite{Caselaw}), and contract provisions (patents, the LED-GAR corpus of contract clauses \cite{tuggener-etal-2020-ledgar}, and a German contracts corpus \cite{niklaus2023multilegalpile}).
\item \textbf{General language}: English literature (Project Gutenberg \cite{gerlach2020standardized}) and Wikipedia.
\item \textbf{GPT-generated text}: English news and legal-style text generated by ChatGPT, prompted under strict legal contexts (e.g.\ “Draft a court opinion” or “Summarize this statute”) to produce outputs as legally formatted as possible, following established data‐collection protocols \cite{chatgpt_paraphrases_dataset}.
\end{itemize}

All texts were preprocessed uniformly: extracting plain text, lowercasing, removing punctuation, and tokenizing into words \cite{springer21}. We segmented each corpus into chunks of roughly \(3\times10^5\) tokens to ensure stable estimates of scaling exponents.

\section{Results}
\vspace{-0.2cm}

In the following subsections, we concentrate on the three primary scale‐free measures: Heaps’ exponent $\beta$, Taylor’s exponent $\alpha$, and gzip compression rate $r$. All of the resulting values are shown in Table ~\ref{tab:result}. The normalized vocabulary entropy $H_{\mathrm{norm}}$ will be revisited in Section~\ref{sec:fred}, where we demonstrate that, alone, it does not reliably distinguish text types.

\begin{figure}[t]
\vspace{-0.3cm}
\centering
\includegraphics[width=0.95\textwidth]{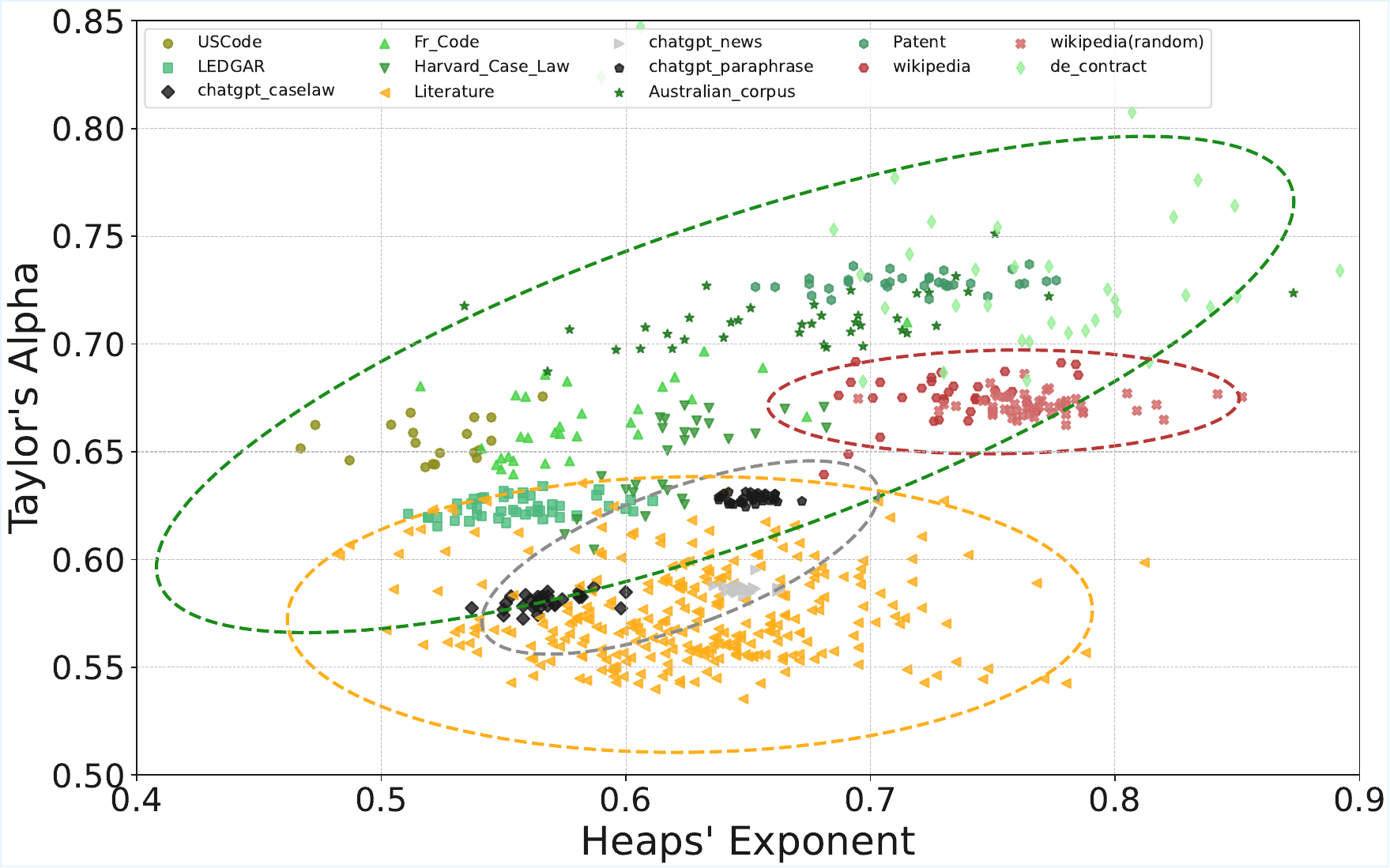}
\caption{Scatterplot of Taylor’s $\alpha$ versus Heaps’ $\beta$ for the selected categories, with legal corpora (green) and GPT outputs (gray) highlighted by ellipses.}
\label{fig:scatter_plot}
\vspace{-0.6cm}
\end{figure}

\begin{figure}[b]
\centering
\footnotesize 
\makebox[\textwidth]{\includegraphics[width=0.85\textwidth]{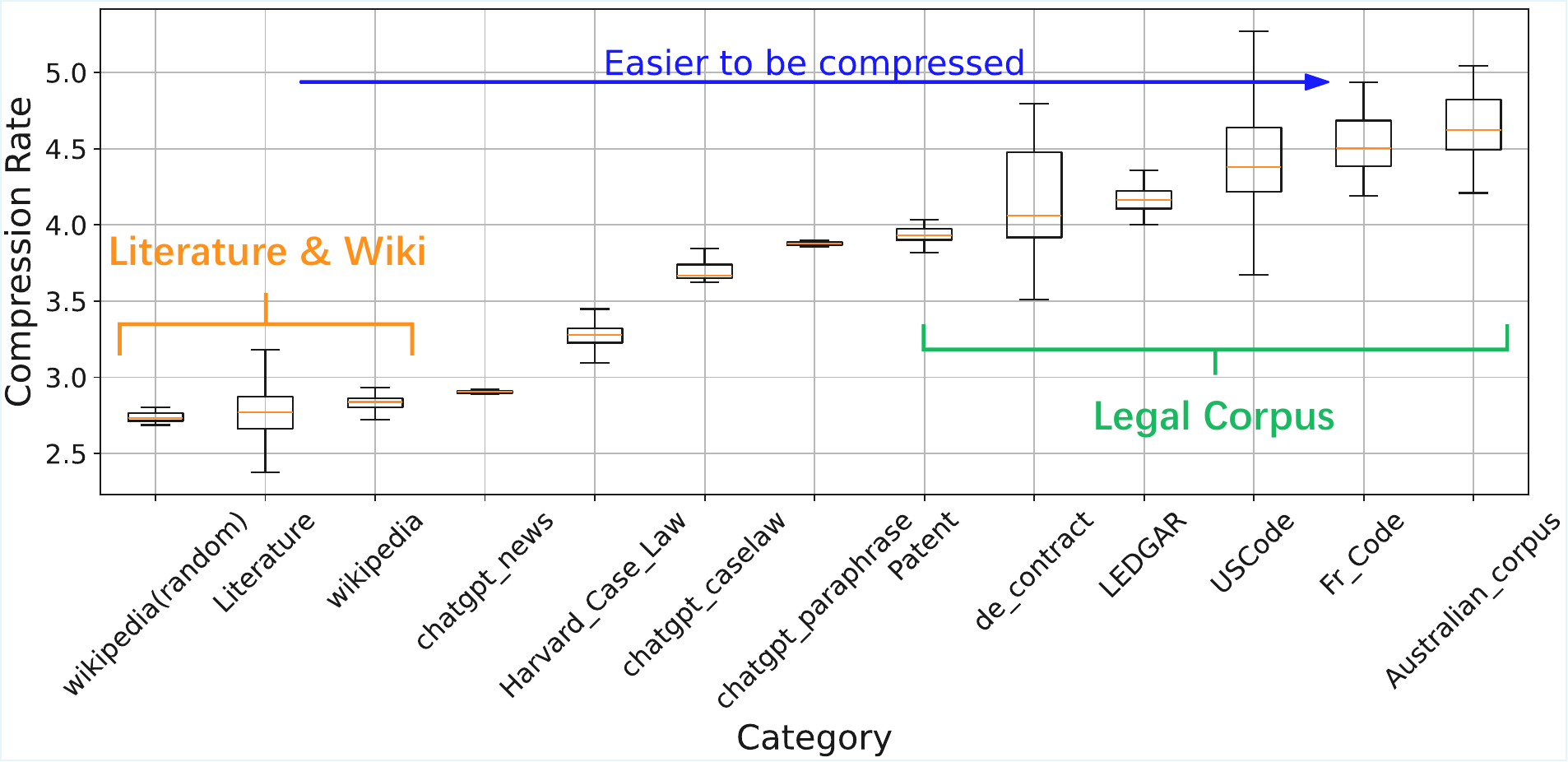}}
\caption{Box plot of compression rate ($r$) per category.}
\label{fig:boxr}
\vspace{-0.9cm}
\end{figure}

\begin{table}[htbp]
\centering
\caption{Average experimental results for different text categories (mean $\pm$ std).}
\begin{tabularx}{\textwidth}{|X|c|c|c|c|}
\hline
Category & \makecell[c]{Heaps' Exponent\\($\beta$)} 
         & \makecell[c]{Taylor's Alpha\\($\alpha$)} 
         & \makecell[c]{Compression Rate\\($r$)} 
         & \makecell[c]{Normalized Entropy\\($H_{\mathrm{norm}}$)} \\
\hline
\multicolumn{5}{|c|}{\textbf{Legal Texts}} \\
\hline
Fr Code (statute)               & 0.584 $\pm$ 0.042 & 0.665 $\pm$ 0.018 & 4.559 $\pm$ 0.322 & 0.782 $\pm$ 0.010 \\
USCode (statute)                & 0.521 $\pm$ 0.025 & 0.655 $\pm$ 0.009 & 4.632 $\pm$ 1.355 & 0.786 $\pm$ 0.040 \\
Australian(statute)     & 0.674 $\pm$ 0.062 & 0.711 $\pm$ 0.012 & 4.684 $\pm$ 0.293 & 0.752 $\pm$ 0.015 \\
\hline
Harvard Case(cases)        & 0.622 $\pm$ 0.026 & 0.647 $\pm$ 0.020 & 3.281 $\pm$ 0.161 & 0.796 $\pm$ 0.014 \\
\hline
Patent (deed)                   & 0.717 $\pm$ 0.032 & 0.729 $\pm$ 0.004 & 3.934 $\pm$ 0.053 & 0.790 $\pm$ 0.008 \\
LEDGAR (deed)                   & 0.556 $\pm$ 0.024 & 0.624 $\pm$ 0.005 & 4.170 $\pm$ 0.094 & 0.765 $\pm$ 0.006 \\
de\_contract (deed)              & 0.769 $\pm$ 0.069 & 0.733 $\pm$ 0.038 & 4.659 $\pm$ 1.578 & 0.785 $\pm$ 0.101 \\
\hline
\multicolumn{5}{|c|}{\textbf{Non-legal Language}} \\
\hline
Literature                      & 0.623 $\pm$ 0.053 & 0.576 $\pm$ 0.023 & 2.634 $\pm$ 0.538 & 0.790 $\pm$ 0.023 \\
wikipedia                       & 0.732 $\pm$ 0.027 & 0.675 $\pm$ 0.011 & 2.834 $\pm$ 0.047 & 0.815 $\pm$ 0.011 \\
wikipedia (random)              & 0.769 $\pm$ 0.026 & 0.672 $\pm$ 0.005 & 2.739 $\pm$ 0.034 & 0.834 $\pm$ 0.011 \\
\hline
\multicolumn{5}{|c|}{\textbf{GPT-Generated}} \\
\hline
news                            & 0.649 $\pm$ 0.007 & 0.587 $\pm$ 0.002 & 2.904 $\pm$ 0.009 & 0.808 $\pm$ 0.002 \\
paraphrase                      & 0.651 $\pm$ 0.008 & 0.628 $\pm$ 0.002 & 3.877 $\pm$ 0.012 & 0.805 $\pm$ 0.003 \\
caselaw                      & 0.567 $\pm$ 0.013 & 0.581 $\pm$ 0.004 & 3.696 $\pm$ 0.058 & 0.785 $\pm$ 0.002 \\
\hline
\end{tabularx}
\label{tab:result}
\vspace{-0.7cm}
\end{table}

\subsection{Cross‐Domain Comparison}
\vspace{-0.2cm}
\label{sec:textalpha}
Figure~\ref{fig:scatter_plot} plots Taylor’s exponent $\alpha$ against Heaps’ exponent $\beta$ for each category, while Figure~\ref{fig:boxr} shows the distribution of compression rates $r$. Together, these measures trace a coherent signature of text complexity.

Legal corpora (green) occupy the upper‐left region of the $\alpha$–$\beta$ plane, combining slow vocabulary growth ($\beta=0.521–0.674$) with strong term clustering ($\alpha=0.624–0.733$), and elevated redundancy ($r=3.281–4.684$). In contrast, literature (orange) lies toward the lower‐right, with higher $\beta=0.623$, lower $\alpha=0.576$, and reduced redundancy ($r=2.634$). 

Wikipedia (red) shifts further right and upward ($\beta=0.732$, $\alpha=0.675$, $r\approx2.83$), reflecting its blend of topical breadth and consistent editorial conventions. Random‐sample Wikipedia moves still further right ($\beta=0.769$, $\alpha=0.672$, $r\approx2.74$), showing greater lexical variety but similar clustering.

GPT‐generated texts (gray) mirror general‐language patterns in both scale-free metrics: $\beta\approx0.65$ and $\alpha\approx0.60$. Notably, ChatGPT points cluster tightly in a compact region of the $\alpha$–$\beta$ plane, indicating limited variance in term‐introduction and clustering behaviors. Their compression rates overlap with legal texts, as shown in Fig.~\ref{fig:boxr}, suggesting that $r$ alone may not suffice for clear separation between AI-generated and human-written texts, which will also be discussed in Sec.\ref{subsec:entropy_compression} . 

The results of $(\beta,\alpha,r)$ reveal that legal drafting conventions impose a distinct complex structure—slow $\beta$, high $\alpha$, high $r$—which general‐purpose and AI‐generated texts do not naturally replicate.

\subsection{Legal Subdomains}

\begin{wrapfigure}{r}{0.65\textwidth}
   \vspace{-0.7cm}
    \centering
    \includegraphics[width=0.63\textwidth]{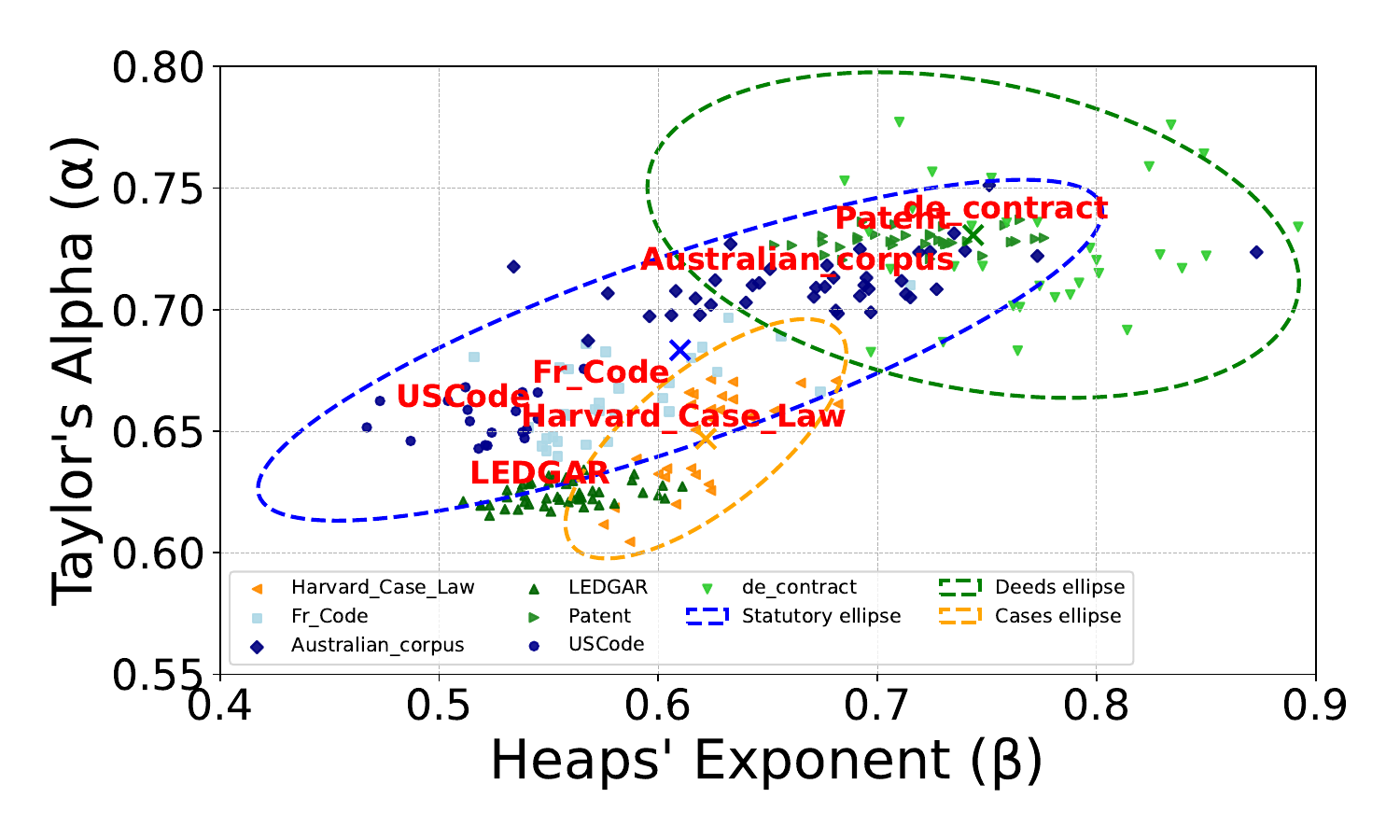}
    \vspace{-0.3cm}
 \caption{Heaps' $\beta$ vs.\ Taylor's $\alpha$ for legal texts.}
    \label{fig:legalscatter}
    \vspace{-0.8cm}
\end{wrapfigure}

Figure~\ref{fig:legalscatter} shows $\alpha$ vs.\ $\beta$ for statutes, cases, and deeds, with compression rates reported in Table~\ref{tab:result} (3rd column).
Statutes (French Code, U.S.\ Code, Australian corpus) cluster at low $\beta$ (0.521  - 0.674), high $\alpha$ (0.655 – 0.711), and the highest $r$ (4.56 – 4.68), reflecting uniform, precise drafting. Case opinions (Harvard Case Law: $\beta=0.622$, $\alpha=0.647$, $r=3.28$) sit slightly lower and on the right, indicating greater lexical flexibility balanced by moderate redundancy.

Deeds split into two groups: patents and German contracts combine rapid term 
growth ($\beta\ge0.717$) with very high clustering ($\alpha\ge0.729$) and redundancy ($r\ge3.93$),\label{sec:legalsub}
while the LEDGAR clause‐level corpus ($\beta=0.556$, $\alpha=0.624$, $r=4.17$) falls nearer to cases, its heterogeneous, single‐clause samples yielding moderate $\alpha$ and redundancy.

These subdomain patterns highlight how scale‐free metrics capture the nuanced drafting conventions across different branches of legal text.
\vspace{-0.3cm}
\subsection{GPT-Generated vs. Human Text}

Finally, we examined texts generated by ChatGPT to rigorously evaluate the linguistic\begin{wrapfigure}{r}{0.65\textwidth}
   \vspace{-0.7cm}
    \centering
\includegraphics[width=0.63\textwidth]{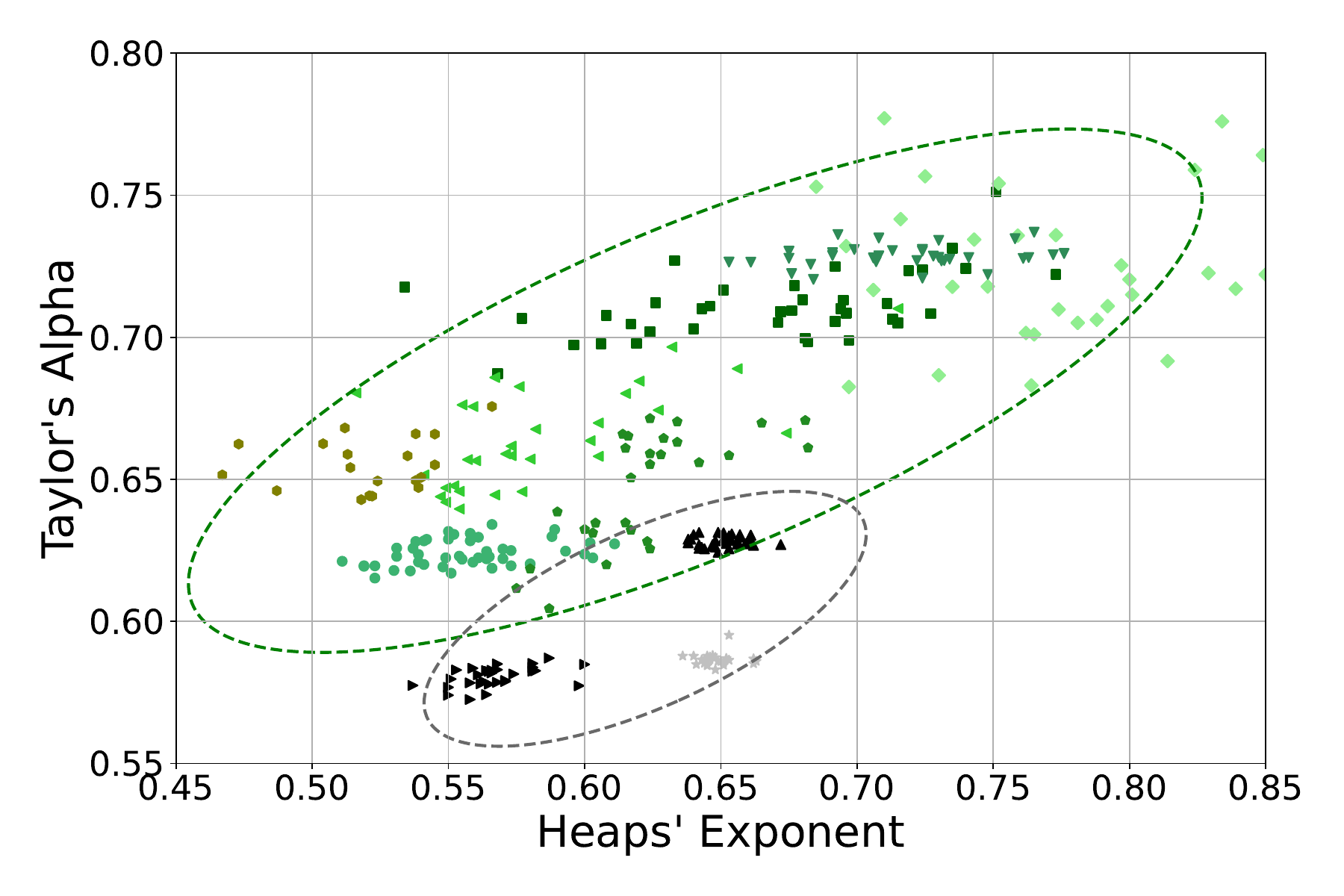}
\vspace{-0.3cm}
 \caption{Heaps' $\beta$ vs.\ Taylor's $\alpha$ for legal (green) and ChatGPT-generated texts (gray \& black).}
\label{fig:legalgpt}
    \vspace{-1.2cm}
  \end{wrapfigure}  characteristics of LLM outputs relative to human-authored legal texts. Figure~\ref{fig:legalgpt} compares legal corpora (green) with ChatGPT outputs (gray) in the $\alpha$–$\beta$ plane (the compression rates are in Table~\ref{tab:result}). GPT‐generated texts shift toward higher $\beta$ and lower $\alpha$, reflecting faster vocabulary turnover and weaker clustering. Moreover, their tight clustering (occupying a small, compact region) signals a lack of variance in generative behavior.  Lower $r$ values further underscore reduced redundancy compared to human‐authored legal drafts.

These empirical observations have significant implications: despite ChatGPT's capability to produce generally coherent and contextually appropriate text, it does not yet replicate the structured linguistic repetitiveness and precision characteristic of true legal drafting. Consequently, these findings suggest that current generative models like ChatGPT may require further refinement or specialized fine-tuning to adequately emulate the statistical signatures intrinsic to legal language.

\vspace{-0.2cm}
\subsection{Normalized Entropy and Compression‐Rate Plane}
\label{sec:fred}
\vspace{-0.2cm}

\begin{figure}[t]
  \centering
  \begin{subfigure}[b]{0.49\textwidth}
    \centering
    \includegraphics[width=\textwidth]{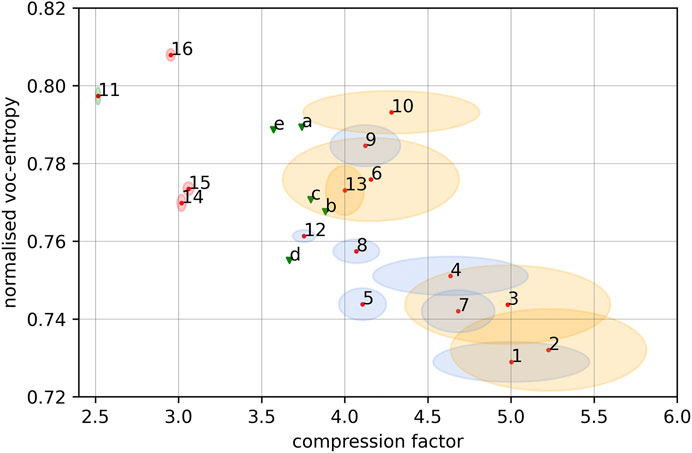}
\caption{Friedrich’s original plane (Compression factor equals to Compression rate). Markers 1–10 and 12–13: legal corpora (acts, regulations, codes). Markers 14–16: EuroParl speeches. Marker 11: literature (Shakespeare). Markers a–e: constitutional texts }
\label{fig:his_plane}
  \end{subfigure}
  \hfill
  \begin{subfigure}[b]{0.49\textwidth}
    \centering
    \vspace{0.12cm}
    \includegraphics[width=\textwidth]{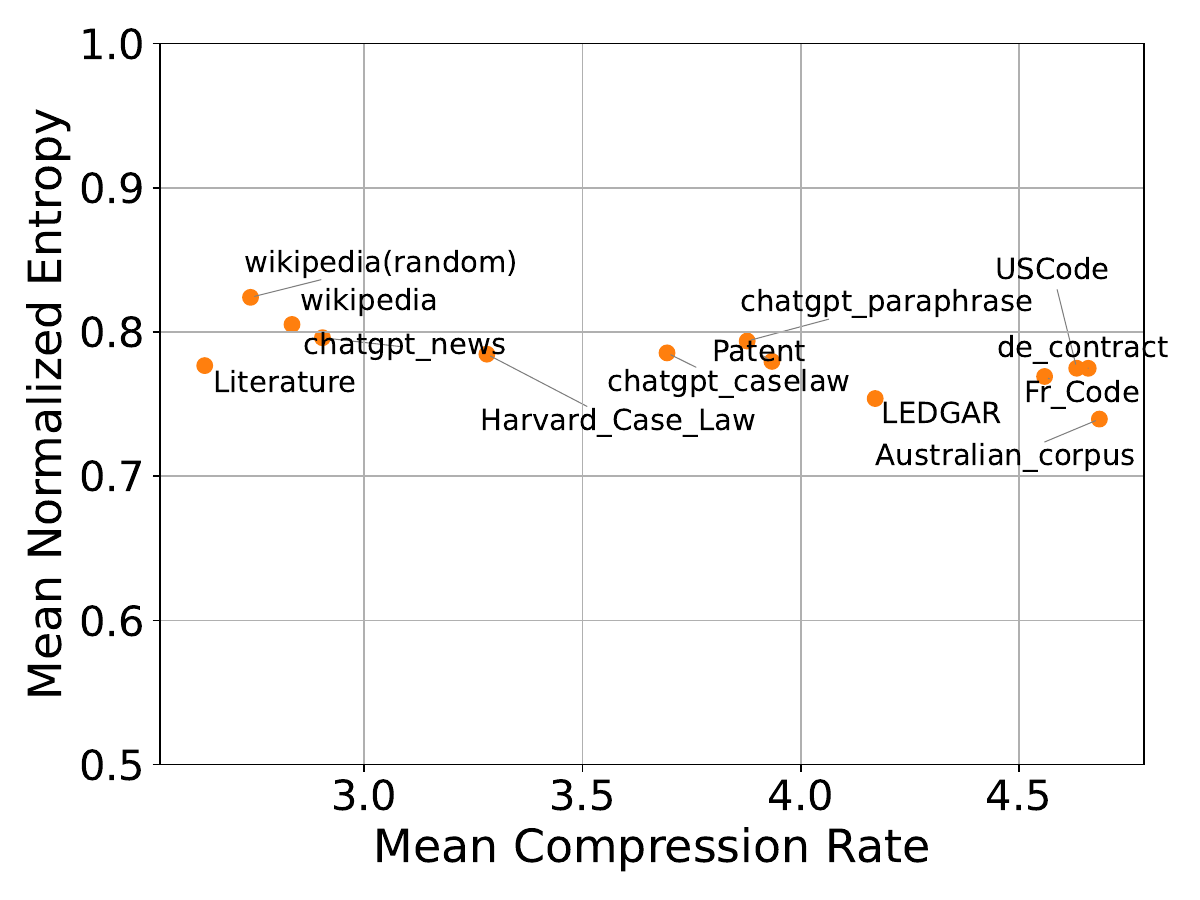}
\caption{Normalized entropy vs.\ compression rate plane with our own results, each point indicates one corpus.}
    \label{fig:our_plane}
    \vspace{0.88cm}
  \end{subfigure}
\caption{Comparison of normalized entropy vs.\ compression rate: (a) Friedrich’s original, (b) with our dataset.}
  \label{fig:entropy_comparison}
  \vspace{-0.7cm}
\end{figure}

\label{subsec:entropy_compression}

One previous work \cite{friedrich2021complexity} employs a two‐dimensional plot of normalized vocabulary entropy \(H_{\mathrm{norm}}\) (See Sec. \ref{subsection3.1}) against gzip compression rate \(r\), suggesting that legal and general‐language texts occupy distinct regions in redundancy space. Crucially, this figure confines \(H_{\mathrm{norm}}\) (vertical axis) to a narrow interval 0.72–0.82, which visually amplifies small corpus‐to‐corpus differences. When we tried to replot our data points (Fig.~\ref{fig:our_plane}), these apparent separations vanished, revealing a single, continuous trend rather than discrete clusters. In other words, \(H_{\mathrm{norm}}\) does not distinguish different kinds of texts by itself.

By contrast, the compression rate \(r\) proves to be a more effective single-axis metric, a conclusion which is consistent with a number of previous studies \cite{entropy16,takahira-etal-2016-upper}. As shown in Fig.~\ref{fig:our_plane} and Fig.~\ref{fig:boxr}, literature and Wikipedia samples occupy the lowest \(r\) values, legal texts occupy the highest, and case opinions and GPT-generated outputs fall in between. Notably, some ChatGPT-generated case-law samples achieve compression rates nearly on par with human-authored statutes, indicating that \(r\) can sometimes overlap across genres—but crucially it still orders the majority of categories in a coherent hierarchy of redundancy.

Despite \(r\)’s discriminative power, our scatterplot of Taylor’s \(\alpha\) versus Heaps’ \(\beta\) (Fig.~\ref{fig:scatter_plot}) reveals an even richer separation. GPT-generated texts form a tight, low-variance cluster in the \((\beta,\alpha)\) plane, as discussed in Sec. \ref{sec:textalpha}, clearly distinct from both literature and legal corpora. This compact clustering highlights GPT’s limited diversity in vocabulary growth and term consistency, whereas combining \(\beta\), \(\alpha\), and \(r\) provides a multidimensional signature that robustly captures domain-specific drafting conventions—something normalized entropy alone cannot achieve.
\vspace{-0.2cm}
\section{Conclusion}
\vspace{-0.2cm}
Our investigation shows that legal language, shaped by precise and repetitive drafting conventions, differs fundamentally from general natural language: it introduces new vocabulary more slowly (lower Heaps’ exponent \(\beta\)), enforces greater term consistency (higher Taylor’s exponent \(\alpha\)), and exhibits higher redundancy (\(r\)) to minimize ambiguity. GPT-generated texts, by contrast, cluster tightly, display more rapid vocabulary growth, and yield reduced redundancy, highlighting that current language models, while proficient in general prose, struggle to replicate true legal writing and lack sufficient variance.

Prior work has shown exceptionally high Taylor’s \(\alpha\) in programming-language corpora \cite{kobayashi-tanaka-ishii-2018-taylors}, suggesting that domain-specific discourse structures drive these complexity signatures. Future efforts should explore fine‐tuning language models to respect such scale‐free patterns—across legal, code, and other specialized genres—to enable AI systems that truly understand and generate text with the appropriate statistical and logical conventions.

\vspace{-0.5cm}
\bibliographystyle{splncs04}
\bibliography{paper}

\renewcommand{\footnotesize}{\scriptsize}
\appendix
\vspace{-0.5cm}
\section{Data Sources and Preprocessing}
\subsection{Data Sources}
\textbf{Legal Texts}: We collected public legal documents from multiple jurisdictions.
\begin{itemize}
\item \textbf{Fr\_Code (France)}: French civil code \cite{niklaus2023multilegalpile}.
\item \textbf{USCode}: United States Code (federal statutes)\footnote{\url{https://uscode.house.gov/}}.
\item \textbf{Australian\_corpus}: Statutes and regulations from an Australian legal corpus \cite{butler-2024-open-australian-legal-corpus}.
\item \textbf{Harvard\_Case\_Law}: A corpus of U.S. judicial opinions \cite{Caselaw}.
\item \textbf{Patent}: U.S. patent texts.\footnote{\url{https://www.uspto.gov/ip-policy/economic-research/research-datasets}}
\item \textbf{LEDGAR}: A contract provision dataset (multi-domain) \cite{tuggener-etal-2020-ledgar}.
\item \textbf{de\_contract}: A corpus of German contract texts \cite{niklaus2023multilegalpile}.
\end{itemize}

\noindent\textbf{General Natural Language Texts}: 

\begin{itemize}
\item \textbf{Literature (Gutenberg)}: A collection of English novels from Gutenberg \cite{gerlach2020standardized}.
\item \textbf{Wikipedia (ordered)}: Selected Wikipedia articles.
\item \textbf{Wikipedia (random)}: Randomly sampled Wikipedia articles.
\end{itemize}

\noindent\textbf{GPT-Generated Texts}: Texts generated by ChatGPT (GPT-3.5) in different tasks:

\begin{itemize}
\item \textbf{chatgpt\_news}: News-style articles written by ChatGPT.
\item \textbf{chatgpt\_paraphrase}: Paraphrasing outputs by ChatGPT.
\item \textbf{chatgpt\_caselaw}: Fictional cases texts generated by ChatGPT.
\end{itemize}

\vspace{-0.5cm}
\subsection{Preprocessing}
All texts were lowercased and tokenized by whitespace. Punctuation was retained. Each corpus was concatenated and then segmented into roughly 300,000-token chunks for analysis. This segmentation preserves context while controlling for sample size~\cite{jpc18}.

\section{Declarations}
\begin{itemize}
\item Funding: This work was supported by JST BOOST, Japan Grant Number JPMJBS2429.
\item Competing interests:The authors have no competing interests to declare that are relevant to the content of this article
\item Ethics approval and consent to participate: Not applicable.
\item Consent for publication: Not applicable.
\item Data availability: The corpora used in this work can be found from in-text citations and links.
\item Materials availability: Not applicable.
\item Code availability: Not applicable.
\item Author contribution: All authors contributed equally to this work.
\end{itemize}
\end{document}